\documentclass[11pt]{article}
\usepackage{coling2020}
\usepackage{times}
\usepackage{url}
\usepackage{latexsym}

\usepackage{booktabs} % For formal tables
\usepackage[normalem]{ulem}
\usepackage{xcolor} %%xl: I need xcolour....
\usepackage{algorithm}
\usepackage{algpseudocode}
\usepackage{amsmath}
\usepackage{mathrsfs}
 \usepackage{amssymb}
\usepackage{subfigure}
\usepackage{makecell}
\usepackage{mathtools}
\usepackage[font=rm]{caption}
\DeclareCaptionType{copyrightbox}
\usepackage{shortvrb}
\usepackage{tabularx}
\usepackage{verbatim}
\usepackage{xspace}
\usepackage{listings}
\usepackage{fontawesome}
\usepackage[multiple]{footmisc}
\usepackage[all]{nowidow}
\usepackage{balance}
\usepackage{microtype}
\usepackage{wrapfig}
\usepackage[medium,compact]{titlesec}
\usepackage{enumitem}
\setlist{itemsep=0pt,parsep=0pt}

\colingfinalcopy

\newcommand{\Patk}[1]{\mbox{\Pat@$#1$}}
\newcommand{\RBPatp}[1]{\mbox{\RBP@$#1$}}

\newcommand{\NDCGatk}[1]{\mbox{\NDCG@$#1$}}
\newcommand{\ERRatk}[1]{\mbox{\ERR@$#1$}}

\newcommand{\argmax}{\operatornamewithlimits{argmax}}

\newcommand{\myurl}[1]{{\url{#1}}}

\newcommand{\myparagraph}[1]{\vspace{0.2\baselineskip}\noindent{\textbf{#1}}.~}

\newcommand{\mycomment}[1]{}

\newlength{\onedigit}
\newcounter{todocount}
\title{Interactive Question Clarification in Dialogue via Reinforcement Learning}

\author{\hspace{-1cm}
  \vspace{0.4cm}
   Xiang Hu\footnotemark[2]    \\
  \hspace{14cm} Ant Financial Services Group\footnotemark[2]\\
  \hspace{14cm} Hasso Plattner Institute, University of Potsdam\footnotemark[3]\\
    \\\And
  \hspace{-1.4cm} Zujie Wen\footnotemark[2] \\
  \\\And
  \hspace{-1.5cm} Yafang Wang \footnotemark[2] \thanks{\ \  corresponding author, email: yafang.wyf@antfin.com}   \\
  \\\And
  \hspace{-1.4cm} Xiaolong Li\footnotemark[2] \\
  \\\And
  \hspace{-1cm}Gerard de Melo\footnotemark[3]  \\
  }

\date{}

\begin{document}
\maketitle
\begin{abstract}
%通过反问的方式澄清可能带来无法预期的用户反馈，在真实系统中引入不确定性。Question reformulation方法往往无法估计所有潜在可能性。因此我们使用一种交互式的问题澄清方法。
% \todo{explain defect of previous works}
Coping with ambiguous questions has been a perennial problem in real-world dialogue systems.
Although clarification by asking questions is a common form of human interaction, it is hard to define appropriate questions to elicit more specific intents from a user.
In this work, we propose a reinforcement model to clarify ambiguous questions by suggesting refinements of the original query. We first formulate a collection partitioning problem to select a set of labels enabling us to distinguish potential unambiguous intents.
We list the chosen labels as intent phrases to the user for further confirmation. The selected label along with the original user query then serves as a refined query, for which a suitable response can more easily be identified.
The model is trained using reinforcement learning with a deep policy network. 
We evaluate our model based on real-world user clicks and demonstrate significant improvements across several different experiments.
% The ability to ask clarification questions to solve ambiguity and missing information phenomena is essential for question answering systems. The current research mainly uses questions generation or questions ranking to ask a clarification question, which lead to low success rate and redundant information. Insufficient use of the graphic user interface (GUI) results in more interactions with users. There is usually no guarantee for replying the user after the clarification. To solve these problems, we propose a question clarification method based on intents recommendation. intents are extracted from the historical Frequently Asked Questions(FAQ) of our system. The recommended intents can provide more concise candidates for user to click. Once an intent is clicked, the system guaranteed to provide a clear question list relative to the real question. We use the reinforcement learning method to recommend intents, and the most challenging problem is cold start. The reward is designed to recommend the most relevant clear question list and maximize the information gain after clicking one intent for better question clarification. The method we proposed for question clarification can solve both ambiguity and missing information phenomena. Experiments on interactions with more than 100 million real-world online users shows the effectiveness of this method. 
\end{abstract}

\section{Introduction}

\blfootnote{
\hspace{-0.65cm} % space normally used by the marker
This work is licensed under a Creative Commons
Attribution 4.0 International License.
License details:
\url{http://creativecommons.org/licenses/by/4.0/}.
}

In real-world dialogue systems, a substantial portion of all user queries are ambiguous ones for which the system is unable to precisely identify the underlying intent. 
%For example, nearly 30\% of user queries in a real-world QA system are ambiguous questions. % Can't give statistics in an academic paper without mentioning details
We observed that many such queries in our question answering (QA) system exhibited one of the following two characteristics.
\begin{enumerate}[partopsep=0pt,topsep=0pt]
    \item Lack of semantic elements such as subject, object, or predicate, e.g. ``How to apply", ``Credit card".
    \item Ambiguous entities, e.g.\ ``My health insurance" (because health insurance consists of numerous sub-categories).
% (3) Misspelling ambiguous. ``how to exist" , where ``exist" may be a misspelling of ``exit". \\
\end{enumerate}
%The ambiguous questions in our QA system can be summarized into 2 types:\\
% (1) \\
% (2) 

\begin{figure}[htb!]
\centering
\vspace{-10pt}
\includegraphics[width=0.61\textwidth]{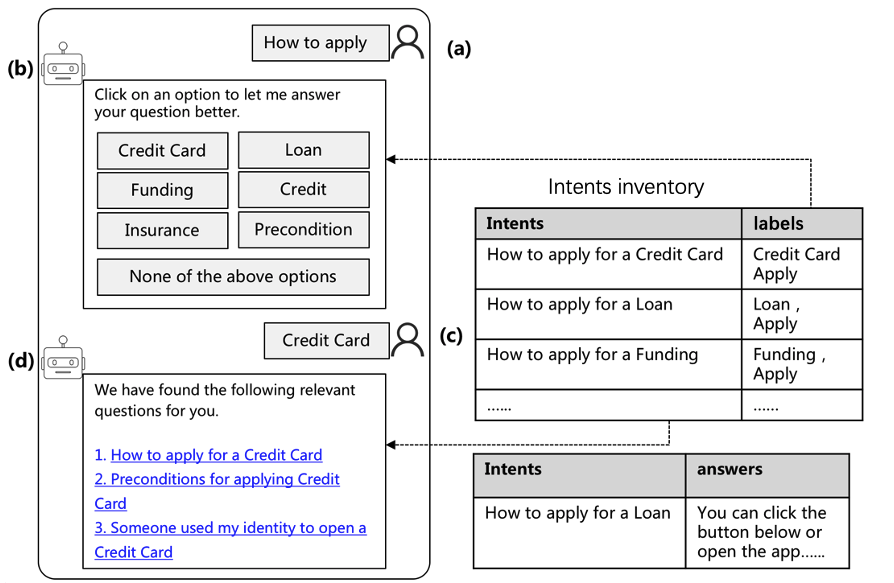}
\caption{Interactive question clarification example. a) The user provides an incomplete or ambiguous question. b) The agent suggests pertinent labels. c) The user confirms by selecting one such label. d) The agent considers the label in conjunction with the original query as a refined query and responds to it.}
\vspace{-2pt}
\label{fig:intro}
\end{figure}

\noindent Given such limited information, it is difficult for a system to accurately respond to a user's ambiguous queries, often resulting in that the user's needs cannot be addressed. For example, the specific intent underlying an utterance such as ``How to apply?" remains obscure, because there are too many products related to the action of ``applying". In practice, one often needs to fall back to human agents to assist with such requests, increasing the workload and cost. The main purpose of deployed automated systems is to reduce the human workload in scenarios such as customer service hotlines. The lack of an ability to deal with ambiguous questions may directly lead to these sessions being transferred to human agents. In our real-world customer service system, this affects up to 30\% of sessions. Hence, it is valuable to find an effective solution to clarify such ambiguous questions automatically, greatly reducing the number of cases requiring human assistance. 

Automated question clarification involves confirming a user's intent through interaction. 
%\cite{DBLP:conf/chiir/RadlinskiC17,DBLP:journals/nle/QuarteroniM09,DBLP:conf/acl/DaumeR18,DBLP:conf/naacl/RaoD19,xu-etal-2019-asking}.
% is essential for a Question Answering (QA) system. 
Previous work has explored asking questions \cite{DBLP:conf/chiir/RadlinskiC17,DBLP:journals/nle/QuarteroniM09,DBLP:conf/acl/DaumeR18,DBLP:conf/naacl/RaoD19}. Unfortunately, clarification by asking questions requires substantial customization for the specific dialogue setting. It is challenging to define appropriate questions to guide users towards providing more accurate information. Coarse questions may leave users confused, while overly specific ones may fail to account for the specific information a user wishes to convey. 

In our work, we thus instead investigate interactive clarification by providing the user with specific choices as options, such as intent options \cite{tang2011towards}.
Unlike previous work, we propose an end-to-end model that suggests labels to clarify ambiguous questions. 
An example of this sort of approach is given in Figure~\ref{fig:intro}.
Here, we consider a closed-domain QA system, where a typical method is to build an intent inventory to address high-frequency requests. In this setting, the set of unambiguous candidate labels for an ambiguous user utterance corresponds to a set of frequently asked questions covered by the intent inventory. 
%By constraining the problem to close-domain, the potential clear questions of an ambiguous question is a finite set. 
In a closed domain, we consider the candidate set to be finite.
For example, in Figure~\ref{fig:intro}, there are three specific intents corresponding to the ambiguous question ``How to apply". 

Our approach induces phrase tags as \emph{labels} for each intent. Thus, we have a catalog of intents with corresponding labels that can be presented to the user. The challenge lies in selecting a suitable list of labels that can effectively clarify the ambiguous question. In our approach, the problem of finding the label sequence is formulated as a collection partitioning problem, where the objective is to cover as many elements as possible while distinguishing elements as clearly as possible. 
% According to Aristotle, \emph{the definition of a species consists of genus proximum and differ}. The differential is the attribute by which one species is distinguished from all others of the same genus. 
The task of question clarification thus amounts to obtaining a suitable set of labels.
\noindent The main contributions of our work are:
\begin{enumerate}{\leftmargin=1em}
\setlength{\itemsep}{0pt}
\setlength{\parsep}{0pt}
\setlength{\parskip}{0pt}
    % \item Propose a novel interactive question clarification method based on sequential intents recommendation.
    \item We formulate interactive clarification as a collection partitioning problem. 
    %Contribute annotated corpora consists of ambiguous questions and potential FAQs pairs, based on the problem definition.
    \item We propose a novel reward function to evaluate the clarification ability of phrase collections and an end-to-end sequential phrase recommendation model trained with reinforcement learning.
    \item  Both offline and online experiments confirm that our method  outperforms pertinent baselines significantly. 
\end{enumerate}

\section{Related Work}
% Question clarification is a long-standing problem, and there are several studies on asking clarification questions. 

\myparagraph{Query Refinement} 
Several works explore the use of clarification questions for query refinement \cite{DBLP:conf/www/KotovZ10,DBLP:conf/coling/SajjadPG12,DBLP:conf/ijcnlp/ZhengSCZ11,DBLP:conf/aaai/MaLK10,DBLP:conf/www/SadikovMWH10}. For instance, \newcite{DBLP:conf/www/KotovZ10} and \newcite{DBLP:conf/coling/SajjadPG12} use question templates to generate a list of clarification questions. \newcite{Elgohary:Peskov:Boyd-Graber-2019} rewrite questions using the dialogue context. \newcite{DBLP:conf/cikm/Zhang0H19} invoke graph edit distance for query refinement. Other studies rely on reinforcement learning to refine user queries \cite{DBLP:conf/emnlp/NogueiraC17,DBLP:conf/iclr/BuckBCGGHW18,DBLP:conf/cikm/LiuZYCY19}, but consider queries that are unambiguous (though possibly ill-formed or non-standard). Accordingly, they seek to increase the recall, while in our setting, we consider ambiguous user queries, and our model primarily seeks to address the task of question clarification.

\myparagraph{Dialogue} \newcite{DBLP:conf/naacl/BoniM03} developed an algorithm to recognize clarification dialogue, rather than for asking clarification questions.
\newcite{DBLP:conf/sigdial/VargesQRI10} found that the use of clarification has a positive effect on concept precision in task-oriented dialogue. \newcite{DBLP:conf/iclr/LiMCRW17a} focus on clarification in the specific circumstance of a bot not understanding a teacher because of spelling mistakes, which is a sub-problem of our setting. \newcite{DBLP:conf/cikm/ZhangCA0C18} generate clarification questions using language patterns with predicted aspect. They do not use reinforcement learning to optimize the order of the questions. \newcite{DBLP:conf/acl/HuangNWL18} devised soft and hard-typed decoders to generate good questions by capturing different roles of different word types. \newcite{DBLP:conf/sigir/AliannejadiZCC19} designed a two-stage retrieval and ranking model to rank clarification question candidates generated by human annotators, different from our end-to-end reinforcement learning approach. \newcite{DBLP:journals/taslp/KorpusikG19} construct clarification questions from a food attribute list (brand, fat, etc.). They rely on a hybrid reinforcement learning approach to select the order of clarification questions to ask, while we present an end-to-end reinforcement learning method. 

\myparagraph{Question Answering} Some studies focus on clarification questions in a community question answering setting \cite{DBLP:conf/chiir/BraslavskiSAD17,DBLP:conf/acl/DaumeR18,DBLP:conf/naacl/RaoD19}. 
% In particular, \cite{Stoyanchev_towardsnatural} construct rules for the generation of targeted clarification questions. \cite{xu-etal-2019-asking} focus on the clarification of entity ambiguity. 
These share in common that they seek to rank or generate clarification questions, while our approach uses reinforcement learning to perform sequential label recommendation for question clarification. The key differences between our work and \newcite{tang2011towards} are three-fold. First, they rely on an ontology, which limits the applicability of their approach in real-world deployments and prevents us from being able to compare against their approach in our experiments, since each domain requires a custom ontology.
%Although our interactive question clarification is similar to~\citet{tang2011towards}, as our domain terms are not in HowNet (Chinese WordNet) which is used for clustering, we omit this method as a baseline.
Second, they cluster the keywords through the ontology, based on templates to achieve a refinement of questions, without using machine learning. Third, they rely on clustering to increase the keyword diversity, while we design a reward with an information gain term that automatically encourages diversity.

\section{Preliminaries}\label{sec:preliminaries}

\begin{wrapfigure}{r}{0.45\textwidth}
  \centering
  \includegraphics[width=0.4\textwidth]{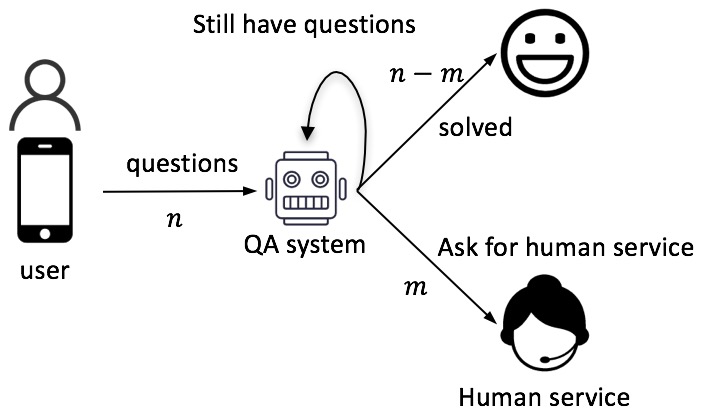}
  \caption{Pipeline of our QA system. $\frac{m}{n}$ is the rate of transferal to human agents (THA). }
\vspace{-10pt}
%   The main purpose is to reduce $\frac{m}{n}$ as much as possible under the premise of ensuring that the user's problem is solved.}
  \label{fig:pipeline}
\end{wrapfigure}

\myparagraph{System overview}
%The rate of transferring to human agents in real application is the core indicator to compare in experiments. 
In order to provide a more concrete picture of our approach, we first briefly describe our QA system, illustrated in Figure~\ref{fig:pipeline}, as an example of how this approach can be instantiated. 

When the conversation exceeds a certain number of rounds or the user explicitly requests human service, the conversation is transferred to a human customer service agent. In this setting, our clarification method chiefly serves to reduce the workload of those human agents. 
%The complete pipeline of our QA system is shown in in Figure~\ref{fig:pipeline}. 
% so the rate of transferring to human agents is one of the most objective metric which can be drawn by statistic:$\frac{\text{sessions transferred to human agents}}{\text{total sessions}}$.
%
% \begin{figure}
%   \centering
%   \subfigure[Pipeline of our QA system. $m/n$ is the rate of transferal to human agents (THA).]{
%     \label{fig:subfig:a} %% label for first subfigure
%     \includegraphics[width=2.5in]{figs/pipeline.png}}
%   \hspace{1in}
%   \subfigure[Example of relationship between phrases and intents. $A$ and $B$ are different intent groups to divide potential FAQs.]{
%     \label{fig:subfig:b} %% label for second subfigure
%     \includegraphics[width=2.5in]{figs/intention_division.png}}
%   \label{fig:subfig} %% label for entire figure
%   \caption{Say something}
% \end{figure}
%
In our real system, there are two stages: label clarification and intent retrieval as illustrated in Figure~\ref{fig:intro}. The label clarification stage provides 6 labels for the user to confirm. Upon selecting one of the suggested labels, the user question is concatenated with the selected label phrase as a new query input. The intent retrieval stage seeks to provide 3 relevant intents for the user to select according to the concatenated query. These additional labels can help clarify and improve the relevance. 
% The system then retrieves $m$ relevant FAQs for the user as question-oriented interface.
% The question-oriented interface seeks to provide $m=3$ relevant FAQs for the user to select. 
% Essentially, intent-oriented interface can be considered as question-oriented with intent clarification.

%对于一个模糊问题，有两种处理流程。question-oriented和intention-oriented. Question-oriented主要通过展示与用户问题最接近的k个问题，受限于对话系统展示空间，一般展示3条。Intention-oriented的方法即本文所讨论的方法，通过相关意图短语供用户确认来消歧,与原问句拼接后在进入question-oriented流程。
% \todo{add the reason why we design our interface as concise phrases, instead of recommending questions?}

% \begin{figure}[!htb]
% % \begin{wrapfigure}{r}{0.45\textwidth}
%   \centering
%   \includegraphics[width=0.4\textwidth]{figs/intention_division.png}
%   \caption{\label{fig:intention_div}Example of relationship between phrases and intents. $A$ and $B$ are different intent groups to divide potential FAQs.}
% % \end{wrapfigure}
% \end{figure}

\myparagraph{Intent and Label Inventory} % FAQ Knowledge Base with Intents
\begin{figure}[!htb]
% \begin{wrapfigure}{r}{0.45\textwidth}
  \centering
  \includegraphics[width=0.4\textwidth]{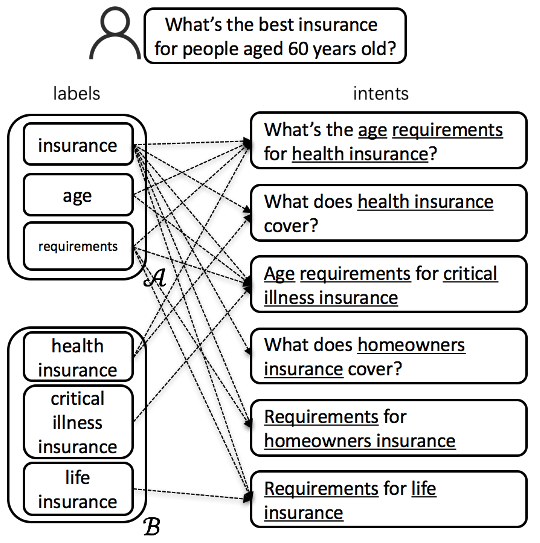}
  \caption{\label{fig:intention_div}Example of relationship between labels and intents. $A$ and $B$ are different label groups to divide potential intents.}
% \end{wrapfigure}
\vspace{-10pt}
\end{figure}
Our system relies on a closed-domain intent and label inventory. 
% FAQs are constructed by clustering users' questions, audited and normalized by experts. Every FAQ has a solution edited by a experts.
The intents along with their corresponding answers are compiled by human experts. The set of labels is a collection of words or phrases that are manually constructed from intents by marking up keywords such as suitable predicates, subjects, or objects. As shown in Figure~\ref{fig:intention_div}, there is a many-to-many relationship between intents and labels. 
% Our task is to find a certain set of intents. The set should recall as many potential FAQs of the ambiguous question as possible, and at the same time, can distinguish potential FAQs from each other.
Note that there is substantial synonymy among the set of labels, which may result in numerous repetitive recommendation results. Thus, ensuring the diversity of the results ought to be a factor in the design of the policy model.

% \todo{duplicate with introduction}

\begin{wraptable}{r}{8cm}
% \begin{table}[!htb]
\centering
 \resizebox{0.45\textwidth}{!}{ %
\begin{tabular}{ll}
\toprule
\textbf{Recall} & \textbf{Valid} \\ \midrule
Can't transfer money using Alpha~~~~ & No \\ %\midrule
\begin{tabular}[c]{@{}l@{}}How to apply for a Credit Card\end{tabular} & Yes \\ %\midrule
\begin{tabular}[c]{@{}l@{}}How to apply for a Loan \end{tabular} & Yes \\ %\midrule
... & ...\\
\bottomrule
\end{tabular}
}
\caption{Example of related intent annotation for user question ``How to apply".}
\label{tb:example}
\vspace{-5pt}
% \end{table}
\end{wraptable}

\myparagraph{Dataset Setup}
\label{sec:data}
In order to solve the cold start problem and evaluate the effectiveness of each model offline, we constructed a benchmark corpus. This annotated corpus consists of 40k ambiguous questions and their potential intents. For this, ten experts were divided into five teams. The two experts in each team annotate the same corpus. Data on which there are disagreements are annotated anew, and only agreed-upon data is selected.
% \todo{data quality}
%
To construct corpora at a relatively low cost, the annotation task is simplified so as to merely elicit a ``yes'' or ``no'' response. The whole annotation process is divided to two stages. At the first stage, we collect ambiguous questions by annotating online query logs. If a query lacks a predicate or the object of the predicate, it is annotated as ambiguous. At the second stage, we annotate potential intents for each ambiguous question. As Table~\ref{tb:example} shows, for each ambiguous question (``How to apply''), the top 50 most relevant intent candidates are collected using the BERT~\cite{DBLP:conf/naacl/DevlinCLT19} semantic similarity model applied to the intent inventory. The human annotators are asked to decide whether an intent can possibly address a user's question. 
% Modeling a ranking model is beyond the scope of this paper.

\section{Reinforcement Learning for Label Recommendation}
\label{sec:methodolody}
\begin{figure}[!htb]
\centering
\includegraphics[scale=0.5]{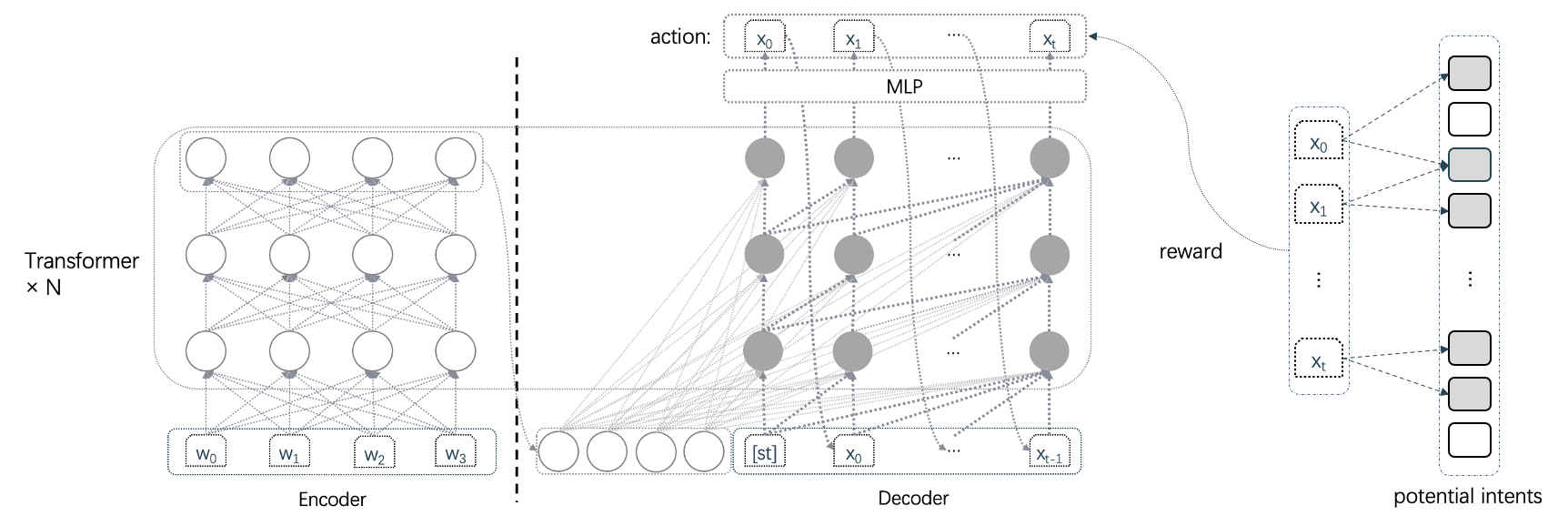}
\vspace{-5pt}
\caption{Label recommendation policy model architecture.}
% Every time a intent is selected, corresponding slot is masked. $\mathcal{Q}(q)$ is human annotated potential intent}
\label{fig:rl_overview}
\vspace{-10pt}
\end{figure}

\myparagraph{Label Recommendation as an RL problem}
In order to train a model able to recommend labels one by one, we have two options: 1) Deduce a path reversely for supervised learning. 2) Create an environment for the model to explore. We believe that creating an environment for the model to explore different label sequences may lead to better generalization ability, which is confirmed in our comparative experiments. We can cast our label recommendation in the reinforcement learning paradigm as in Figure \ref{fig:rl_overview}. Our model can be viewed as an \emph{agent} that interacts with an \emph{environment}, which consists of the user question and recommended labels. The action space consists of more than 1,000 candidate labels, out of which a suitable next label needs to be selected as a next action. In order to increase the diversity and reduce the number of synonymous labels, our model takes historical recommended labels into account. Upon having recommended $N$ labels, the final reward (introduced later) is assigned and the parameters are updated.
%为了提升多样性，降低同义意图，我们为每个意图创建独立的状态向量，他们会随着已推荐的标签而改变。所有意图集合的向量构成了强化学习的状态向量。

\myparagraph{Policy Model} As $N$ labels to be recommended could be considered as a sequence, we use a seq2seq architecture to model the problem. As shown in Figure \ref{fig:rl_overview}, in the encoder stage, the query is encoded by BERT and a vector representation is generated. In the decoder stage, the input at time step $t$ is the action at step $t-1$ (step 0 is [st]). For each step, one-way multi-head attention \cite{DBLP:conf/nips/VaswaniSPUJGKP17in} is applied on previously recommended labels and the vector representation of the input query. Finally, the action probability at each step is estimated.

\myparagraph{Rewards} Intuitively, the chosen labels ought to maximize the recall of the intents with regard to the human-annotated potential intents. However, a trajectory with high recall may not be sufficient for clarification, as high recall can easily be achieved by suggesting labels such as in group $A$ in Figure \ref{fig:intention_div}. Rather, a good label set should efficiently discriminate between potential intents as in group $B$ in Figure~\ref{fig:intention_div}. We recast this as a collection partition problem. Subsequently, inspired by the ID3 algorithm \cite{DBLP:journals/ml/Quinlan86}, we use Information Gain as a term to evaluate the final reward. 

Formally, given a user query $q$, and the human-annotated potential intents $\mathcal{Q}(q)$, 
%the total count of $\mathcal{Q}(q)$ is $|\mathcal{Q}(q)|$, 
our policy model selects a list of labels $\tau_{N} = \{x_1, x_2, \dots, x_N\}$. We map all the chosen labels $\tau$ to the retrieved potential intent set $S(\tau)$ with a many-to-many relationship between labels and intents: % as defined in Equation~\ref{eq:standard_qset}.
\vspace{-5pt}
\begin{equation}
\small
\label{eq:standard_qset}
\mathcal{S}(\tau)=\bigcup_{x \in \tau} \big[ \mathcal{M}(x)\cap \mathcal{Q}(q) \big]
\vspace{-5pt}
\end{equation}
$\mathcal{M}(x)$ denotes the intent set mapped from label $x$. 
%The total count of $S(\tau)$ is $|S(\tau)|$. 
$\mathcal{K}$ denotes the universe set of intents. An indicator vector $\mathbf{I}(q)=(\mathbf{I}_1,\mathbf{I}_2, \dots, \mathbf{I}_{|\mathcal{K}|})$ indicates for each intent $s^{i}$ in $\mathcal{K}$ whether it exists in the human-annotated intent set $\mathcal{Q}(q)$, as defined below.
\vspace{-5pt}
\begin{align}
\small
\mathbf{I}_{i}=\left\{\begin{matrix}
1 & s^{i} \in \mathcal{Q}(q)\\ 
0 & s^{i} \notin \mathcal{Q}(q)
\end{matrix}\right.
\vspace{-5pt}
\end{align}
The probability that an intent is the answer to an ambiguous question is computed as
%Equation~\ref{eq:standard_prob}.
\vspace{-5pt}
\begin{equation}
\small
\label{eq:standard_prob}
P(s^{i} \mid q)=\frac{\mathbf{I_i}}{|\mathcal{Q}(q)|}.
\vspace{-5pt}
\end{equation}

\noindent We define potential intents recalled at time step $t$ as $S(\tau_{t})$, the conditional entropy of $S(\tau_{N})$ is $\mathcal{H}(\tau_{N})$, defined
as follows.
%in Equation~\ref{eq:ent_k}
\begin{equation}
\small
\label{eq:ent_k}
\begin{split}
&\mathcal{D}(x_t)=\mathcal{M}(x_t) \cap \mathcal{Q}(q) \setminus \mathcal{S}(\tau_{t-1}) \\
&\widetilde{P}(s \mid q,\tau_{t})=\frac{P(s \mid q)}{\sum\limits_{s' \in \mathcal{D}(x_{t})}^{ }P(s' \mid q)} \\
&\mathcal{H}(x_t)=-\sum_{s\in \mathcal{D}(x_t)}\widetilde{P}(s \mid q,\tau_{t})\,\log{\widetilde{P}(s \mid q,\tau_{t})} \\
\end{split}
\end{equation}
Here, $\mathcal{M}(x_t)$ denotes the set of intents mapped from label $x_t$. $\mathcal{D}(x_t)$ is the marginal recall over the potential intent set $\mathcal{Q}(q)$ for label $x_t$. $\widetilde{P}(s\mid q,\tau_{t})$ is the normalized probability of $P(s\mid q)$ for intents in $\mathcal{D}(x_t)$. 
The entropy at time step 0 is $\mathcal{H}_{0}$, defined as
%According to Aristotle, \emph{the definition of a species consists of genus proximum and differ}. The differential is the attribute by which one species is distinguished from all others of the same genus. It's the same for the question clarification. A good interactive intents set should distinguish potential questions well. 
\vspace{-2pt}
\begin{equation}
\small
\label{eq:ent_0}
\mathcal{H}_{0}=-\sum_{s \in \mathcal{Q}(q)}P(s\mid q) \,\log P(s \mid q).
\vspace{-5pt}
\end{equation}

\noindent The Information Gain is defined as
\vspace{-5pt}
\begin{equation}
\small
\label{eq:division_ent}
\Delta(\tau_{N})=\sum_{t=1}^{N}\frac{\left |\mathcal{D}(x_i) \right |}{\left | \mathcal{S}(\tau_{N}) \right |}\mathcal{H}(x_t)-\mathcal{H}_{0},
\vspace{-5pt}
\end{equation}

%\begin{equation}
%\begin{split}
%\label{eq:division_ent}
%&\mathcal{D}(L^{a_{t}})=\mathcal{M}(L^{a_{t}}) - \mathcal{S}(\tau_{t-1})
%\\&\mathcal{H}(\tau_{0})=-\sum_{s \in \mathcal{S}(q)}^{ }P_{s}(s|q)log P_{s}(s|q)
%\\&\mathcal{H}(\tau_{k})=\sum_{i=1}^{k}\frac{\left \|\mathcal{D}(L^{a_{i}}) \right \|}{\left \| \mathcal{S}(\tau_{k}) \right %\|}\mathcal{H}(L^{a_{i}})
%\\&\mathcal{H}(L^{a_{t}})=-\sum_{s\in \mathcal{D}(L^{a_{t}})}^{ }\widetilde{P}(s|q)\log{\widetilde{P}(s|q)}
%\\&\Delta (\tau_{k})=\mathcal{H}(\tau_{k})-\mathcal{H}(\tau_{0})
%\end{split}
%\end{equation}

\noindent and the final reward is then defined as
%in Equation~\ref{eq:reward_final}
\vspace{-5pt}
\begin{equation}
\small
\label{eq:reward_final}
R(\tau_{N})=\sum_{s \in S({\tau_{N}})}\emph{P}(s\mid q) +\beta\Delta(\tau_{N}).
\vspace{-5pt}
\end{equation}
% The more diverse of marginal potential questions $x_t$ recalls at time step t, the bigger the $\mathcal{H}(x_t)$ and the ambiguity is. 
In our experiments, $\beta$ by default is set to 1.

Considering there are more than 1000 candidate labels, the size of the search space in MCTS may explode. To reduce its size, we only sample labels in $\left \{ x|\mathcal{M}(x)\bigcap\mathcal{Q}(q)\neq \varnothing \right \}$ because only such labels have a relationship with candidate intents worth exploring. Thus, the size of the search space is drastically reduced.

\begin{wrapfigure}{r}{0.5\textwidth}
% \begin{figure}[!htb]
  \centering
  \includegraphics[width=0.5\textwidth]{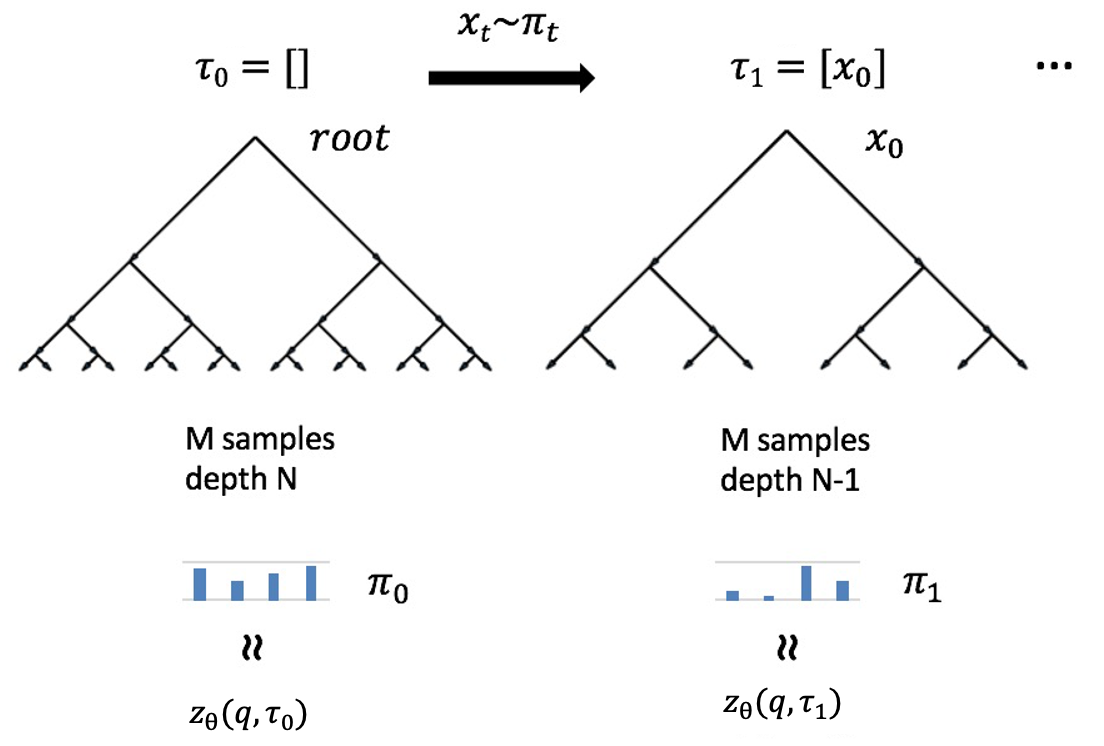}
  \caption{MCTS. At time step $t$, the sampling process keeps searching until it reaches depth $N-t$.}
  \label{fig:MCTS}
  \vspace{-10pt}
% \end{figure}
\end{wrapfigure}

\myparagraph{Training} The policy model to suggest labels is trained from samples generated via a Monte-Carlo tree search (MCTS)~\cite{DBLP:conf/cg/Coulom06,DBLP:conf/ecml/KocsisS06,DBLP:journals/tciaig/BrownePWLCRTPSC12}. The MCTS starts from an empty label set and stops when the trajectory includes $N$ labels, as in Figure \ref{fig:MCTS}.
Each simulation starts from the root state and iteratively selects a move with maximal $V(\cdot)$, which is computed according to the upper confidence bound for tree search \cite{DBLP:conf/ecml/KocsisS06} as
\begin{equation}
\small
\label{eq:UCB}
    % V(v) = \frac{Q(v)}{N(v)}+ c\sqrt[ ]{\frac{2\ln N(v.parent)}{N(v)}}
V(v) = \frac{Q(v)}{N(v)}+ \beta_\mathrm{T}\, \sqrt[ ]{\frac{2\ln N(p_v)}{N(v)}},
\end{equation}
where $p_v$ denotes the parent of $v$ and $\beta_\mathrm{T}$ by default is set to 1. After a path has been sampled, the $Q$ value of each node in the path is updated according to %Equation~\ref{eq:q_update}:
\begin{equation}
\small
\label{eq:q_update}
Q(v)=\frac{\displaystyle\sum_{\tau\in T(v)}^{ }R(\tau)}{N(v)}
% Q(v)=\frac{\sum_\limits{\tau\in T(v)}R(\tau)}{N(v)},
\end{equation}

\noindent where $N(v)$ denotes the visiting time of $v$ and $T(v)$ denotes the set of all trajectories containing $v$. Once the search is complete after $M$ samples, probabilities $\pi$ for the next action are estimated following Equation~\ref{eq:pi_gen}, where $N(\cdot)$ is the visit count of each move from the root state and $T$ is a parameter controlling the temperature.
\vspace{-5pt}
\begin{equation}
\small
\label{eq:pi_gen}
\pi(\cdot \mid v)=\frac{N(\cdot )^{1/T}}{\sum\limits_{v'\in C_v}^{ }N(v')^{1/T}}
\end{equation}
Here, $C_v$ denotes the children of node $v$. Additional exploration is achieved by adding Dirichlet noise $\mathrm{Dir}(\cdot)$ to the prior probabilities as 
%Equation~\ref{eq:random_action} like 
in AlphaZero \cite{SilverMastering}:
\begin{equation}
\small
\label{eq:random_action}
P(\cdot |v)\sim \frac{3}{4}\pi(\cdot \mid v) + \frac{1}{4}\mathrm{Dir}(0.03)
\end{equation}
$x_{t}$ is selected in a weighted round robin manner in accordance with $P(\cdot \mid v)$. The neural network $\emph{z}_{\theta}(q,\tau_{t})$ is adjusted to minimize the KL divergence $D_\mathrm{KL}$
%\footnote{\url{https://en.wikipedia.org/wiki/Kullback\%E2\%80\%93Leibler\_divergence\#cite\_note-1}}
of the neural network estimated probabilities to the search probabilities $\pi$ as:
% In order to reduce the sampling space and the computational complexity of training, we usually truncate the most relevant intents according to the question distribution together with some randomly selected intents as available actions in the training.
\vspace{-10pt}
\begin{equation}
\small
\label{eq:loss_fn}
\mathcal{L}(\tau_N) = \sum_{t=1}^{N}D_\mathrm{KL}\big[\mathbf{\emph{z}_{\theta}}(\cdot|q,\tau_{t})\,||\,\pi(\cdot \mid v)\big].
\vspace{-5pt}
\end{equation}

\section{Experiments}
\vspace{-1pt}
Following standard practices in industry, we first conduct offline experiments to select reasonable models for which we subsequently perform an online evaluation. Only the best-performing model in the online tests is kept running online. We also perform an ablation study on the pipeline without label clarification.
% We compared two kinds of interfaces: question-oriented and intent-oriented ones (cf.\ Section~\ref{sec:preliminaries}).
%
In order to verify whether the Information Gain can help to reduce the overlap between intents and the user question,
% we also design indicators to compare the gap between different methods.
we also perform experiments to evaluate the diversity and complementarity of the label recommendation method.
\subsection{Experimental Settings}
\vspace{-1pt}
\label{sec:settings}
We first conduct offline experiments by using the 40k annotated ambiguous questions and their potential intents as explained in Section~\ref{sec:data}. The corpora are divided into training and test sets at a $9:1$ ratio. 
%
% We compared two kinds of interfaces: question oriented and intent oriented. 
% Question oriented interface is to provide relevant questions for user to click. intent oriented interface is what we proposed in this paper, which provides some intents for user to click before returning relevant questions. These additional intents can help clarify and improve the relevance.
%
% In order to verify whether the information gain can help to reduce semantic overlap between intents and the user question, we also design indicators to compare the gap between different methods.
%
The parameters of our policy model are as follows.
The sample count in MCTS is $M=1,000$. We output $N=6$ intents for each ambiguous question. The total number of training epochs is $E=5$. We use a 12-layer pretrained BERT base model as the encoder for queries and the hyperparameters of the decoder are the same as for the encoder.
%为了验证加入信息增益的方法是否能够有效降低与文本语义重复的问题，我们设计了相关观察指标，用以对比不同方法的差别。
% We also evaluate complementary of different methods.
% \todo{add complementary experiments motivation}
%  To verify diversity and complementary, we designed a series of observational experiments to validate our conjectures. 
 
\subsection{Evaluation Metrics}
\label{sec:metrics}
\myparagraph{Evaluation metrics for offline experiments}
The goal of our offline experiments is to evaluate the label recommendation methods, and select the most promising ones to perform online experiments. 
We evaluate them in terms of \textbf{Recall@N}, which reflects how many intents among all potential intents of $q$ are retrieved among the $N$ intents emitted by the model. 

The key desideratum for label recommendation models is to cover as many potential questions as possible. It is relatively fair to compare the recall of potential intents recommended by different methods on the annotated data set. For label trajectory $\tau_{N}$, the recall can be computed as
%by the formula: 
% \todo{add explanation for M and S}
\vspace{-3pt}
\begin{equation}
\small
\text{recall}(q, \tau_{N})=\frac{\sum_{x \in \tau_{N}}^{ } \left | \mathcal{M}(x)\bigcap \mathcal{Q}(q)  \right |}{\left | \mathcal{Q}(q) \right |}
\vspace{-3pt}
\end{equation}
where $\mathcal{Q}(q)$ is the set of potential intents for ambiguous query $q$, and $\mathcal{M}(x)$ is the set of all intents mapped to intent $x$ in the intent inventory. The upper bound is calculated inversely from the results of annotated corpora:
\vspace{-3pt}
\begin{equation}
\small
\tau_{N}^{*}(q)=\argmax\limits_{\tau_{N}}\sum_{x \in \tau_{N}}^{ }\left | \mathcal{M}(x) \cap \mathcal{Q}(q)  \right |
\vspace{-3pt}
\end{equation}

\noindent $\tau^{*}_{N}(q)$ denotes the set of $N$ best labels covering the potential intents. Thus, the upper bound recall of $q$ would be $\text{recall}(\tau^{*}_{N}(q)$). 

\myparagraph{Evaluation metrics for online experiments}
In our subsequent online experiments, our key metrics are the rate of transferal to human agents (THA) and the click through rate (CTR). In our experiments, every time a question is classified as an ambiguous question, six labels are provided to the user, who may select one of them or just ignore the selection. Given $t$ as the number of times we output labels, and $c$ as the number of times the user selected one of them, we define $\mathrm{CTR} = \frac{c}{t}$. Note that the user may opt to 
%instead select $\emptyset$, i.e., the option of selection 
select \emph{none of the above options}. In this case, the pipeline equals intent retrieval without clarification. The CTR reflects how useful the recommended labels are to users.
% In this case, the intent-oriented interface equals the question-oriented interface.

\myparagraph{Evaluation metrics for complementary experiments}
We compare the repetition rate at the word piece level of labels generated by two methods as an experiment to evaluate the diversity. The diversity is quantified as:
\vspace{-3pt}
\begin{equation}
\small
    \mathrm{div}(\tau_{N})=\frac{\left | \mathcal{W}(\tau_{N}) \right |}{\displaystyle\sum_{w \in \mathcal{W}(\tau_{N})}^{ }{C}(w)},
\vspace{-3pt}
\end{equation}
where $\mathcal{W}(\tau_{N})$ is the set of word pieces tokenized from the labels, and ${C}(w)$ denotes the number of times word piece $w$ appears among the labels. We also count the overlap rate:
\vspace{-3pt}
\begin{equation}
\small
    \mathrm{overlap}(\tau_{N}, q)=\frac{\displaystyle\sum_{x \in \tau_{N}}^{ }\sum_{t \in \mathcal{T}(x)\bigcap \mathcal{T}(q)}^{ }C(t)}{\displaystyle\sum_{x \in \tau_{N}}^{ }\sum_{t \in \mathcal{T}(x)}^{ }C(t)}
\vspace{-3pt}
\end{equation}
Here, $\mathcal{T}(x_{t})$, $\mathcal{T}(q)$ denote the tokens sets of $x_{t}$ and $q$, respectively. The overlap thus essentially reflects the number of tokens of labels appearing in a query. 

% \todo{add CTR in our experiments}
%CTR是在给用户推荐6个标签后，用户点选6个中的一个的比例。(他可以选择以上都不是或继续对话无视标签)

% \todo{explain offline and online experiments}
%我们实验主要分为两部分，线上实验和线下实验。线上实验通过用户真实反馈对比不同方案的差异。通过控制变量法对比用户使用不同服务后的转人工率及点击率的差异。我们在同一时间段进行横向对比，确保用户问题的分布一致。在所有其他条件都相同的情况下，只有对模糊问题的处理方法有区别。线下实验无法直接衡量解决问题的能力，可以通过设计一系列基础指标，侧面观测不同方法差异性。

\subsection{Baselines}

Several methods for label clarification serve as baselines for the offline experiments, while our method is denoted as \textit{RL (ours)}. 
% In addition to comparing phrase recommendation methods, the online experiment performed an ablation experiment by directly returning k most similar intents without providing options for clarification.
%Question-oriented methods are only used in the online evaluation.
%线上实验除了对比intent-clarification的方法，还进行了ablation实验，移除意图消歧环节观察数据差异。
%解释对比的原因，本质上是ablation实验，本文的消歧流程是消歧再top-k intents,因此对比没有消歧直接top-k intent用以观察引入消歧带来的提升。
%\myparagraph{Relevance ranker} Given a query, similarity scores with candidate intents are evaluated by a semantic similarity model. Considering the display limitation of the dialog robot, 3 intents will be presented. The model is a 12-layer BERT which takes concatenation of two sentences as input and is trained by the corpus of customer service.
% \subsubsection{Intent-oriented methods}
\subsubsection{Label Clarification Methods}
% 给定一个q，通过q与潜在intent集合的标注数据训练一个分类模型。通过分类模型得到q对应潜在候选intent的概率分布。通过规则的方法，找到一个意图集合，可以使其覆盖的intent集合的概率只和最大。规则方法如公式所述:
% \myparagraph{PMI+IDF} Method introduced in \cite{tang2011towards} \\
% %问题定义及解决方法有较大区别，该方法主要解决开放域问题，本文直接参考其实验结果
\myparagraph{Supervised} Given a query and a set of potential intents, there are limited labels related to the potential intents set. Traverse all possible label sequences over the limited labels set and choose the one with the highest rewards as the ground truth. If there are multiple sequences corresponding to the highest reward, pick one randomly.

\myparagraph{Greedy} Given a user question, we train a classification model on the annotated corpus of ambiguous questions and the corresponding potential intents by minimizing the loss function
\vspace{-5pt}
\begin{equation}
\small
\label{eq:kl_loss}
\mathcal{L}=\sum_{q}^{ }D_\mathrm{KL}[f_{\theta}(\cdot | q)\parallel P(\cdot |q)]
\vspace{-5pt}
\end{equation}
The classification model $f_{\theta}$ is used to estimate the probability distribution $P(\cdot |q)$ of the potential intents. Through this greedy method, our goal is to find a set of intents for which the sum of the probabilities of intents they cover is as high as possible. The greedy rule is given by $\mathrm{Score}(x_t)=\sum\limits_{s \in \mathcal{D}(x_t)}f_{\theta}(s, q)$, where
%
% \todo{not clear} Given a user question, similarity scores of candidate intents with similarity model. Considering the display limitation of the dialog robot, 6 intents will be presented. Take the highest score of intents over marginal coverage set $D(x_t)$ at each step:
%
%\myparagraph{Rule based} Given a query, get similarity scores of candidate intents with similarity model. Considering the display limitation of the dialog robot, 6 intents will be presented. Take the highest score of intents over marginal coverage set $\mathcal{D}(L^{a_{t}})$ at each step:
% \vspace{-3pt}
% \begin{equation}
% \small
% \mathrm{Score}(x_t)=\sum\limits_{s \in \mathcal{D}(x_t)}f_{\theta}(s, q)
% % \vspace{-3pt}
% \label{eq:rule_score}
% \end{equation}
$\mathcal{D}(x_t)$ is the marginal recall of intents described in Section~\ref{sec:methodolody}. At each time step $t$, we select the label with the highest score as $x_{t}$. Thus, the label set is generated by the rule.

\myparagraph{RL (no state transition)} As another baseline, we explore the implication of not taking recommended labels into account. This is a BERT classification model which outputs the intent with the highest probability at time step $t$ and masks it at the next time step.
% Given a query, return six intents by no state transition model. It is a simple BERT classification model which outputs a intent with highest probability at time step t with intents in $\tau_{t-1}$ masked.
% \todo{what is no state transition model}.

% \myparagraph{RL (seq2seq)} Given a query $\mathbf{q}$, it returns $k$ intents ($k=6$) by LSTM based seq2seq model \todo{add reference}. 

% Though our interactive question clarification is similar to , the intent phrases in our domain is not included in Hownet (Chinese WordNet) and methodologies differs a lot and evaluate metrics, so it's not included in baselines.

% Although our interactive question clarification is similar to~\citet{tang2011towards}, as our domain terms are not in HowNet (Chinese WordNet) which is used for clustering, we omit this method as a baseline.
%it is not compared as baseline.
%虽然\cite{Tang}的问题与我们相似，但由于我们领域的专业术语不在hownet中，聚类效果较差，并且方法与评价指标区别较大，因此不在对比实验中

\subsubsection{Ablation Study}

\myparagraph{Top-K intents} To contrast the truncated interface with the original, full interface, we retrieve the $m$ most similar intents in terms of semantic similarity without interacting with users. The detail of intent retrieval is described below. (Note that for a label-oriented interface, after the user selects one label, the original query is concatenated with the label phrase as a new query and relevant intents are retrieved by the same model.)

\myparagraph{Intent retrieval} For each query, a list of potential intents can be retrieved and ranked by BM25. We re-rank the candidates by applying BERT model to estimate the semantic similarity between query and each candidate. The model is a 12-layer BERT, which takes the concatenation of two sentences as input. Considering the display limitation of the dialogue bot environment, the top three results are presented to the user.

% this method also returns to the related intent

%
\subsection{Results}
\myparagraph{Offline experiments} The experimental results in Table~\ref{tb:offline} show that our method significantly outperforms others. 
% Meanwhile, the coverage rate of our reward is close to the result of single coverage reward. It is worth noting that the model based on LSTM decoder doesn't perform well. The reason is that decoder changes the current state of the query without explicitly changing the intents state. And we explicitly apply recommended intents to rests to better model the mutual exclusion of similar intents to improve diversity and coverage. \todo{not clear}
The greedy method has limited recall due to its reliance on the accuracy of its classification model. We observed that it is difficult to achieve a satisfactory recall by estimating potential intent probabilities through a classification model. 

\begin{wraptable}{r}{7cm}
% \begin{table}[htb!]
\centering
\resizebox{0.45\textwidth}{!}{ %
\begin{tabular}{lll}
\toprule
 & \textbf{labels=3} & \textbf{labels=6} \\ \midrule
Greedy & 19.47\% & 32.01\% \\
Supervised & 45.72\% & 51.53\% \\
% RL(seq2seq) & ?\% \\
% PMI+IDF & 56\% \\
RL (no state transition) & 17.23\% & 29.83\% \\
% RL(lstm) & 39.31\% & 45.15\% \\
RL (ours) & \textbf{52.45\%} & \textbf{57.22\%} \\
\midrule
Upper bound & 60.46\% & 67.34\% \\
\bottomrule
\end{tabular}
}
 \vspace{-3pt}
\caption{\label{tb:offline}Offline experimental results.}
 \vspace{-9pt}
% \end{table}
\end{wraptable}
%
% \begin{wraptable}{r}{7cm}
% % \begin{table}[htb!]
% \centering
%  \resizebox{0.45\textwidth}{!}{ %
% \begin{tabular}{lll}
% \toprule
%  & \textbf{Recall@3} & \textbf{Recall@6} \\ \midrule
% RL (no state transition) & 17.23\% & 29.83\% \\
% Rule based & 19.47\% & 32.01\% \\
% % RL(seq2seq) & ?\% \\
% % PMI+IDF & 56\% \\
% RL(recall) & \textbf{54.76\%} & \textbf{59.72\%} \\
% RL(ours) & 52.34\% & 57.22\% \\ 
% \midrule
% Upper bound & 69.46\% & 70.34\% \\
% \bottomrule
% \end{tabular}
% }
%  \vspace{-3pt}
% \caption{\label{tb:offline}Offline experimental results.}
%  \vspace{-9pt}
% % \end{table}
% \end{wraptable}

%And diversity relies on rule-based clustering which has poor adaptability across domain. % ????
%The main reason is that the recall of the rule-based method relies too much on the accuracy of the classification model. In experiments, 
Our policy model also significantly outperforms the model without state transitions, confirming the need for considering the action history. The labels recommended by simple classification models do not yield sufficient diversity, resulting in very low recall. By modeling the problem as a seq2seq one, our model learns to recommend a next label that differs from previous ones, thereby improving the recall of potential intents. 

It is worth noting that the supervised method outperforms all other baselines except ours. We believe that it does not explore the training data sufficiently. In most cases, there are multiple label sequences that can get similar rewards, and the supervised method can only consider one of them as the ground truth, remaining unable to explore equally good or second-best paths, which leads to insufficient exploration of labels. Thus, the search of the supervised method is not as exhaustive as our method's. Our results are close to the theoretical upper bound, which is further corroborates the effectiveness of our method.
%We shall compare these two choices in our online experiments.
% 基于规则的方法的召回结果过于依赖分类模型的准确性，通过实验对比可以发现靠分类模型直接拟合潜在intent分布很难达到满意的召回。相比于不带意图状态迁移的，我们的策略模型显然高出很多。该实验验证了Policy Model对意图状态建模的有效性。简单的分类模型推荐的标签不具备多样性，导致recall很低。通过对每个意图建立状态向量，可以训练模型学习推荐与之前结果差异性较大的意图，从而提升潜在标问的召回。可以看到，我们的结果已经比较接近理论上界，也论证了policy model的有效性。虽然RL(c)在召回上高于RL(c+e)，但差距不明显，需要通过线上实验对比真实世界的效果。

\begin{wraptable}{r}{5.5cm}
\small
\centering
 \resizebox{0.35\textwidth}{!}{ %
\begin{tabular}{lll}
\toprule
 & \textbf{THA} & \textbf{CTR} \\
 \midrule
Top-K intents& 15.40\% & \multicolumn{1}{c}{-} \\
RL (recall) & 14.51\% & 62.61\% \\
RL (ours) & \textbf{14.20\%} & \textbf{66.36\%} \\
\bottomrule
\end{tabular}
}
\caption{\label{tb:online_exp}Online experimental results.}
% \end{table}
\vspace{-10pt}
\end{wraptable}

\myparagraph{Online experiments}
The offline experimental results show that \textit{RL (no state transition)} and the \textit{Greedy} method do not perform well, leaving only \textit{RL (ours)} for the online experiments. Here we mainly compare the performance of two rewards: recall only and reward + entropy.
We compare label recommendation methods and perform an ablation study using real online user clicks.
% We compare question-oriented methods and intent-oriented methods in our real system. 
For this, we collected data over a period of two weeks in our real deployment.
The experimental results, illustrated in Table~\ref{tb:online_exp}, show that the CTR of \textit{RL (ours)} is significantly higher than for \textit{RL (recall)}. We believe that this gap objectively reflects the importance of entropy to improve the quality of the label set. Furthermore, \textit{RL (ours)} also outperforms \textit{RL (recall)} with regard to the rate of transferal to human agents (THA). The Top-k intents method directly retrieves the most relevant three questions without interacting with users. The THA gap between Top-K intents and RL based methods reflects the contribution of label clarification. The experiments show that our method has a positive effect with regard to the system's ability to clarify ambiguous questions, reducing the workload of human agents.

%Top-k intents方法直接索引最相关的3个问题,intent clarification方法在与用户交互后，将原query与用户选择的phrase拼接，再进行top-k intents索引。因此本质上是一种ablation对比，从THA的差距可以客观反映intent clarification带来的贡献。

% \begin{table}[htb!]

% \begin{figure}[htb]
%   \centering
%   \includegraphics[width=0.5\textwidth]{figs/comparison_curve.png}
  
%   \caption{ \label{fig:comparison_curve}Rate of transferal to human agents and CTR.}
%   \vspace{-8pt}
% \end{figure}

% \begin{figure}[htb]
%   \centering
%   \includegraphics[width=0.5\textwidth]{figs/man_calling_curve.png}
  
%   \caption{Rates of transferring to human agents in two weeks.}
%   \label{fig:man_calling_curve}
% \end{figure}
% \begin{figure}[htb!]
%     \centering
%     \includegraphics[width=0.5\textwidth]{figs/CTR.png}
%     \caption{Click through rates of intents in two weeks.}
%     \label{fig:ctr}
% \end{figure}

% The way of intent interaction is closer to the dialogue of human being. In particular, such approach can cover the various possibilities of potential intents and clarify user's question. Therefore, compared with the relevant ranker, our method has obvious advantages on reducing the workload of human agents. \todo{add some words before to be fluent with previous paragraph}

% The experiment results show that CTR of $RL (e+c)$ is significantly higher than the $RL (c)$. We believe that this gap objectively reflects the importance of entropy to improve the quality of intents set. Furthermore, $RL(e+c)$ also outperforms $RL(c)$ in rate of transferring to human agents. 

\subsection{Complementary Evaluation}

\begin{table}[!htb]
\centering
 \resizebox{0.4\textwidth}{!}{ %
\begin{tabular}{ll}
\toprule
\multicolumn{2}{l}{How to apply} \\ \hline
\multicolumn{1}{l|}{RL (recall)} & apply, register, credit card \\ \hline
\multicolumn{1}{l|}{RL (ours)} & credit card, loan, QR code \\ \midrule
\multicolumn{2}{l}{How to claim insurance?} \\ \hline
\multicolumn{1}{l|}{RL (recall)} & \begin{tabular}[c]{@{}l@{}}claim, health insurance,\\ medical insurance\end{tabular} \\ \hline
\multicolumn{1}{l|}{RL (ours)} & \begin{tabular}[c]{@{}l@{}}health insurance, medical insurance,\\ homeowners insurance\end{tabular} \\ \midrule
\multicolumn{2}{l}{What was the payment just now?} \\ \hline
\multicolumn{1}{l|}{RL (recall)} & \begin{tabular}[c]{@{}l@{}}transaction records, inquire,\\ transfer money\end{tabular} \\ \hline
\multicolumn{1}{l|}{RL (ours)} & \begin{tabular}[c]{@{}l@{}}billing details, transition records,\\ inquire records\end{tabular}\\
\bottomrule
\end{tabular}
}
\vspace{-3pt}
\caption{Excerpts of outputs from different methods. For simplicity, only the first three are displayed.\label{tb:showcase}}
\vspace{-3pt}
\end{table}

By inspecting specific cases, we find that the main difference between \textit{RL (recall)} and \textit{RL (ours)} is the complementarity with the user's question. Taking ``How to apply" in Table~\ref{tb:showcase} as an example, \textit{RL (recall)} selects ``apply", ``register", which exhibit semantic overlap with the question itself. Though these may lead to improved recall of potential intents, they do not enable any further clarification. The results of \textit{RL (ours)} include products that one can apply for, helping to establish the user's underlying intent. For a recall-only approach, the labels that yield the highest rewards must be the ones with the highest semantic overlap. Hence, it is inevitable that repetitive information will be chosen, thereby making a part of the label set redundant.

\begin{wraptable}{r}{7cm}
% \begin{table}[htb!]
\centering
% \resizebox{0.37\textwidth}{!}{ %
\begin{tabular}{llll}
\toprule
 & \textbf{Diversity} & \textbf{Overlap} & \\ \midrule
Greedy & 75.27\% & 6.39\% \\
RL (recall) & 79.92\% & 9.69\% \\
RL (ours) & \textbf{80.10}\% & \textbf{7.69}\% \\
\toprule
\end{tabular}
%}
% \vspace{-2pt}
\caption{\label{tb:complementary}Complementarity evaluation: The lower the overlap, the better the complementarity.}
%\vspace{-5pt}
% \end{table}
\end{wraptable}

To verify our conjecture, we compare the diversity and complementarity using the indicators introduced in Section~\ref{sec:metrics}. Although the two indicators are not precise metrics for diversity and semantic overlap, they help to assess the gap of the models trained with the two different reward mechanisms.
As we can see from Table~\ref{tb:complementary}, the reinforcement learning methods significantly surpass the \textit{Greedy} method on diversity, but the two RL methods are comparable to each other. This illustrates that recall as a reward is a major contribution to diversity. On its own, the overlap indicator is not meaningful, as it can be reduced to $0$ by recommending irrelevant labels. But along with the recall, the difference in overlapping rate illustrates the effectiveness on reducing semantic repetition. 
Therefore, the proposed reward is superior to all other compared methods.

\section{Conclusion}

We present an end-to-end model to resolve ambiguous questions in dialogue by clarifying them using label suggestions. We cast the question clarification problem as a collection partition problem. In order to improve the quality of the interactive labels as well as reduce the semantic overlap of the labels and the user's question, we propose a novel reward based on recall of potential intents and information gain. We establish its effectiveness in a series of experiments, which suggest that this novel notion of clarification may as well be adopted for other kinds of disambiguation problems.
%Our experiments shows that the way of intent interaction is more effective in solving user problems than returning relevant results. At the same time, through the comparison of online ctr, it fully proves that the intents recommend by the policy model trained via our new reward is more helpful to users.

\bibliographystyle{coling}
\bibliography{main}

\end{document}